# Dataset Creation Pipeline for Camera-Based Heart Rate Estimation

Mohamed Moustafa[1,2], Amr Elrasad[2], Joseph Lemley[2],

Peter Corcoran[1,2]

[1]University of Galway, Galway, Ireland;

[2]Xperi Corporation, Galway, Ireland;

## ABSTRACT

Heart rate is one of the most vital health metrics which can be utilized to investigate and gain intuitions into various human physiological and psychological information. Estimating heart rate without the constraints of contact-based sensors thus presents itself as a very attractive field of research as it enables well-being monitoring in a wider variety of scenarios. Consequently, various techniques for camera-based heart rate estimation have been developed ranging from classical image processing to convoluted deep learning models and architectures. At the heart of such research efforts lies health and visual data acquisition, cleaning, transformation, and annotation. In this paper, we discuss how to prepare data for the task of developing or testing an algorithm or machine learning model for heart rate estimation from images of facial regions. The data prepared is to include camera frames as well as sensor readings from an electrocardiograph sensor. The proposed pipeline is divided into four main steps, namely removal of faulty data, frame and electrocardiograph timestamp dejittering, signal denoising and filtering, and frame annotation creation. Our main contributions are a novel technique of eliminating jitter from health sensor and camera timestamps and a method to accurately time align both visual frame and electrocardiogram sensor data which is also applicable to other sensor types.

## 1. INTRODUCTION

Heart rate is considered one of the most vital metrics within the context of health monitoring. Heart rate is influenced by a myriad of factors such as drugs [1], change in bodily temperature [2], cognitive state [3], and fitness levels. Monitoring heart rate can give insights into lifespan and metabolic syndromes [4], risk of a heart attack [5], risk of sudden death [6], emotions [7], and social behaviour [8].

Correspondingly, heart rate estimation is of great relevance even outside hospitals and medical institutions, where the reliance on contact-based sensors becomes either too inconvenient or even dangerous (e.g. sensors could distract a car driver and slow down their reaction). Monitoring an individual's heart rate during certain frequent activities such as driving a car [9] or working in risky environments [10] is of particular importance.

Camera-based heart rate estimation, where a visual sensor (usually a camera) is aimed at the subject and used to estimate their heart rate, presents itself as a convenient and reliable alternative to classical contact-based heart rate estimation outside the context of urgent health care (e.g. while driving, working, or visiting a store) [11]. As a result, research and implementation of visual heart rate estimation has seen various improvements [12,13] and has expanded to include data from not just visible light cameras, but even near-infrared [14] and thermal [15] cameras.

Crucial for such improvements was better access to higher volumes of detailed heart rate datasets [16–18] on which new techniques could be trained, tuned, and tested before being deployed. This is especially true for machine and deep learning methods where large volumes of data can be fully leveraged to exceed performance of classical techniques while handling variations in environment and subject pose much better than traditional image processing methods [12]. However, whether researchers wish to rely on publicly accessible data or collect their own, knowledge of dataset preparation and data processing is of dire importance. Unfortunately, published work usually either excludes certain data preparation steps or only mention data preparation steps necessary for their specific algorithmic implementation while skimming through any non-novel steps.

In this work, we propose a new approach to create an accurately aligned heart rate dataset from acquired visual and health sensors. Our method includes time alignment and de-jittering of the collected samples' timestamps which is skipped by most researchers. We also propose a novel technique of eliminating jitter from health sensor and camera timestamps and a method to accurately time align both visual frame and Electrocardiogram (ECG) sensor data. Additionally, our work is not only suitable for heart signal dataset creation but can be implemented in other applications where a sensor with a known sampling rate is used. To the best of our knowledge, our proposed algorithm and pipeline have not been published before.

The rest of this paper is organised as follows: Section 2 explores related literature. Sections 3 and 4 describe our proposed pipeline and the experimental results. Section 5 presents concluding remarks.

## 2. RELATED WORK

It is more common to find data preparation details within papers whose main contribution is either a new dataset or estimation methodology than to find publications entirely focused on the data preparation pipeline. Even then, the details of time-alignment and data synchronisation are often skipped-over or insufficiently discussed.

[19] details the undertaken data preparation steps as one of the main contributions. ECG readings are converted to an RRI signal using the Pan-Tompkins algorithm [20] followed by processing to detect and remove ectopic beats, abnormal beats caused by unusual impulses. After that, the RRI signal is detrended using a wavelet method then resampled using cubic spline interpolation.

In [21], a video sequence is decomposed into temporally consistent superpixels from which a tentative remote photoplethysmography (rPPG) signal is extracted. The signal is then further processed as to enhance physiological information extracted from skin regions by calculating the signal-to-noise ratio (SNR) relying on Fourier transforms and double-step functions for the first and second harmonics. A weighted average based on the SNR is then calculated to reconstruct the rPPG signal.

In [22], the authors propose a Kalman filter-based method to solve the time synchronization problem of asynchronous measurement from multiple sensors. This is further developed in [23, 24] to account for software delays and dropped measurements.

Within the context of data mining data preparation, a systematic mapping study (SMS) [25] was conducted into the use of preprocessing techniques in clinical datasets which analyzed and classified more than 100 papers. The SMS concluded that the most recurrent preprocessing tasks in medical disciplines were data reduction and data cleaning.

## 3. DATA PREPARATION

In this section we go through various tasks involved with the preparation of a dataset for the task of camera-based heart rate estimation.

### 3.1 Elimination of Faulty Data

After data acquisition, the first step taken should be to ensure that the data has been saved properly without any corruption. Following that, the data is to be reviewed, possibly manually, to ensure that it is of satisfactory quality. When estimating heart rate from face videos, the face should be at least partially visible in all frames. It is not uncommon, especially in long, multi-camera set ups, for subjects to move outside the field of view even when instructed not to move. Certain health sensors, such as the pulse oximetry (SpO2) sensor, might not be able to read data properly when moved or jostled from its previous position. In some cases, the resulting signal would not simply have higher noise, but might lack sufficient information to extract heart rate from. The detection and elimination of such data sections is an important step in creating a reliable dataset as otherwise it introduces undue noise into the data which is detrimental when the algorithm aims to leverage trends in the input data to extract or predict information.

### 3.2 De-jittering Timestamps

As frames and sensor readings eventually need to be synchronized to annotate the frames, timestamps play a significant role in dataset preparation. If the data is out of sync, improper labels might be given to frames leading to the model being trained and evaluated incorrectly.

During data acquisition, certain factors such as equipment heating, sensor-to-computer communication channel, usage of lab/company-written software for acquisition (usually to capture data simultaneously from sensors and cameras from different manufacturers) result in fluctuations in sampling and capture rate, dropped data, as well as adding slight delays between data capture and timestamp generation (when timestamps are not generated by the sensor itself but by the software). Reducing such fluctuations serves to improve the accuracy of annotating frames with sensor labels, we refer to this process as timestamp de-jittering.

Our method of timestamp de-jittering, shown in Fig. 1, is based on mapping captured timestamps to synthetically generated, evenly spaced timestamps. This method is very sensitive to sampling rate and therefore caution should be taken that it is calculated to a high degree of accuracy without relying on assumptions (e.g. advertised device sampling rate).

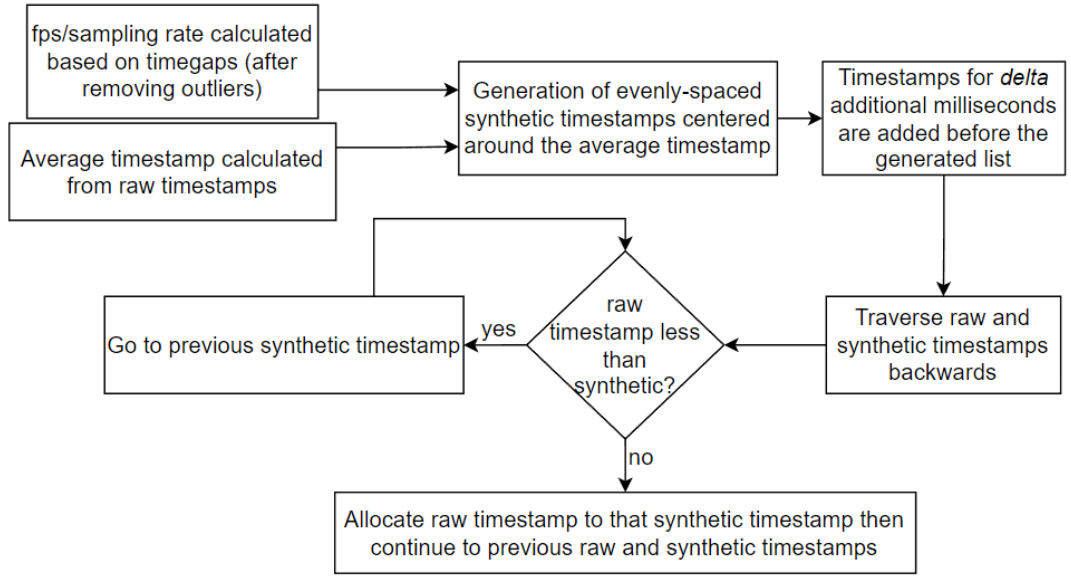

Figure 1. Diagram illustrating overall process of timestamp de-jittering. The process concludes when all captured timestamps have been allocated.

To start with, the data sampling rate is calculated by getting the average timestamp difference (i.e. the temporal gap between each consecutive timestamp), henceforth referred to as timestep, after excluding any timestep outliers lying beyond *m* standard deviations. The sampling rate is one divided by the mean timestep and is usually slightly below the advertised camera or sensor frame/sampling rate (e.g. 29.9fps calculated for a 30fps camera).

Next, synthetic timestamps are generated by first calculating the average timestamp, then generating timestamps before and after the average timestamp. This is done by continuously subtracting and adding the mean timestep calculated previously to the mean timestamp, thus covering the time period originally covered by the captured raw timestamps with evenly spaced synthetic timestamps. The synthetic timestamps cover an additional *delta* millisecond before the start of the captured timestamps to make room for delays to be removed from the timestamps at the start of the acquisition (i.e. if the first timestamp was delayed, correcting it should map it to a previous point in time).

Mapping captured timestamps to synthetic timestamps involves traversing the captured timestamps from end to start. For each raw timestamp, a mapping is created between it and the first non-allocated synthetic timestamp with an equal or lower value. This is because it is not possible for the correct value of a timestamp to be more than the recorded value, which includes the delay and jitter we wish to remove. Expanding the time period covered by the synthetic timestamps serves to compensate for the delay that might have been present in the first timestamp (meaning that synthetic timestamps cannot start from the first recorded timestamp).

### 3.3 Signal Processing

In order to better extract overall trends and information from the captured health sensor recordings, the captured signals go through several processing algorithms as illustrated in Fig. 5.

Firstly, smoothing is applied to reduce noise and emphasize overall peaks. For this purpose, we use the Savitzky-Golay filter, which fits a polynomial function to each window of points size *w*. We found that using a window of size 101 and fitting a quadratic function gave satisfactory results for our use case, this is further discussed in the experimental results.

Next, we apply a bandpass Butterworth filter on the data to get a periodic wave which will be used for annotating the frames as ground truth. Since we wish to filter out noise from the heart rate signal, we use cut off frequencies [0.7, 2.5] Hz based on average human heart rate range from literature [12].

### 3.4 Annotation Creation

As health sensors tend to have higher sampling rate than camera frame rate, downsampling is needed to create labels for each frame. This is done through the usage of timestamps, as shown in Fig. 2.

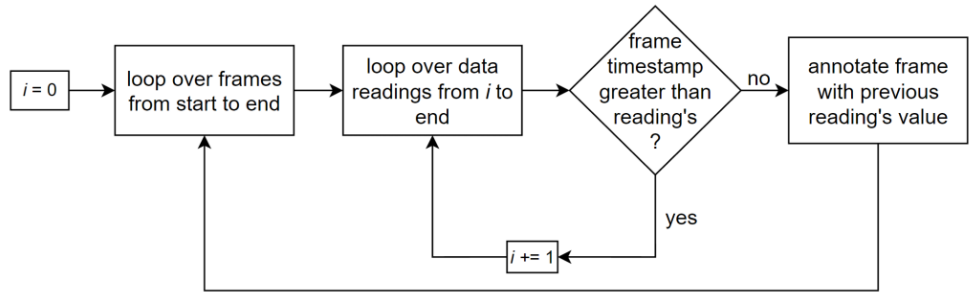

Figure 2. Outline of frame annotation

For each frame, the ECG signal with the greatest timestamp value which is smaller than the frame timestamp value is used to annotate that frame. This method is based on two assumptions: the reading cannot happen after the timestamp and the temporally closest sensor reading represents the physiological frame information most accurately. A global variable was used to store the index of the last traversed ECG reading which lowered the algorithm's time complexity from $O(n^2)$ to $O(n)$.

## 4. EXPERIMENTAL RESULTS

This section aims to illustrate the results obtained by applying the aforementioned data preparation steps on an internal dataset acquired using a 30 frames per second camera and a 1kHz ECG sensor. In our acquired data, the mean and standard deviation of the frame timegaps were 33.33 ms and 11.52 ms and of the ECG reading timegaps were 1.00 ms and 2.63 ms respectively. These results are obtained using m=3, excluding timesteps 3 standard deviations from the mean. After the removal of any faulty data, such as noisy signal sections (as visually observed) or frames where the subject moved out of field of view, the timestamps obtained from the camera and sensors are de-jittered as to more reliably temporally align camera and sensor data.

This is done by mapping the collected timestamps to synthetically generated timestamps. However, not all synthetic timestamps are allocated captured timestamps (i.e. not all synthetic timestamps are used). These instances imply that the readings or frames corresponding to them were dropped (i.e. not being sampled or captured), which appear as timegap peaks as shown in Fig. 3.

Dropped frames or readings are a result of various factors such as the acquisition system not being able to handle the incoming data stream, the hardware not being able to keep up, or some other form of bottleneck in the experimental setup which includes the sensor/camera, connection (wireless or over physical connection), acquisition computer, and the acquisition software.

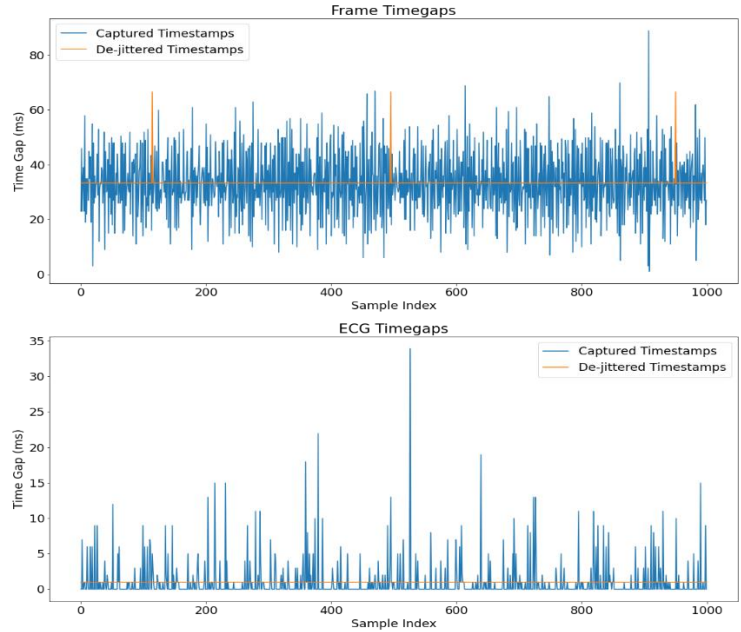

Figure 3. Captured and de-jittered frame timesteps (Top) and captured and de-jittered ECG timesteps (Bottom) samples

Applying our method and the Kalman filter-based method presented in [23,24] on our data, our method reduced the standard deviation of the frame timegaps to 1.64 ms. Meanwhile, the Kalman filter method increased the mean to 54 ms while reducing the standard deviation to 5.47 ms. When applied on ECG timegaps, our method reduces the standard deviation to 0.005 ms while the Kalman filter method increases the mean to 1.62 ms while reducing the standard deviation to 2.05 ms. Our method results in a non-zero standard deviation due to the presence of dropped frames/readings in the data. This feature of our method enables easy identification of points in time where a reading or frame has been dropped. Table 1 below compares the results of the two algorithms against the raw timestamps' mean and standard deviation while Fig. 4 compares the outputs on a sub-sample of data.

|  | Frame Timestamps | | ECG Timestamps | |
| --- | --- | --- | --- | --- |
|  | Mean (ms) | Std (ms) | Mean (ms) | Std (ms) |
| Raw | 33.33 | 11.517 | 1.00 | 2.629 |
| Kalman-filter method | 53.92 | 5.470 | 1.62 | 2.053 |
| **Our method** | **33.33** | **1.642** | **1.00** | **0.005** |
| Theoretical optimal values | 33.33 | 0.000 | 1.00 | 0.000 |

Table 1. Experimental result comparison between Kalman filter-based algorithm and proposed algorithm

As shown, the Kalman filter approach does not work well with highly variable timegap jitter. As illustrated, their method is affected heavily by higher timegap jitter and is not suitable for filtering out/correcting timegap jitter to bring timestamps closer to the expected value. In their paper, the overall jitter in the data used to illustrate their method performance appears to be around 0.2 ms which is a significantly lower jitter than in our experimental data.

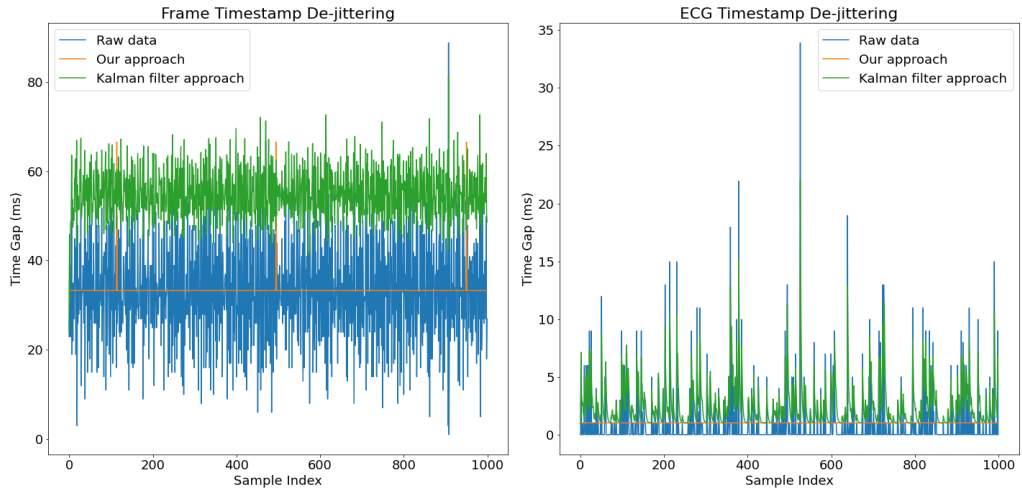

Figure 4. Comparison between our approach and Kalman filter approach on frame and ECG data sample

Following that, smoothing is applied to the ECG signal to remove noise while retaining main trends. We found the fitting a quadratic function to every 101 data points yielded the best results. The window size of 101 was chosen after experimenting with various other widths. A larger window would result into too much smoothing, which would degrade the main peaks present in the data, while a smaller one would not be sufficient to remove noise. However, if the sensor sampling rate is changed, a different window size must be selected.

After smoothing, a bandpass filter with cut-off frequencies [0.7Hz - 2.5Hz] is used to obtain the final label values, shown in Fig. 5, which are to be mapped onto the camera frames. This frequency range is commonly used in heart signal literature. Finally, as downsampling is required due to camera frame rate usually being lower than sensor sampling rate, care needs to be taken as to properly temporally align the downsampled sensor data with the high sampling rate processed readings. A quick way to validate that is to plot both signals in the time domain, as shown in Fig. 6.

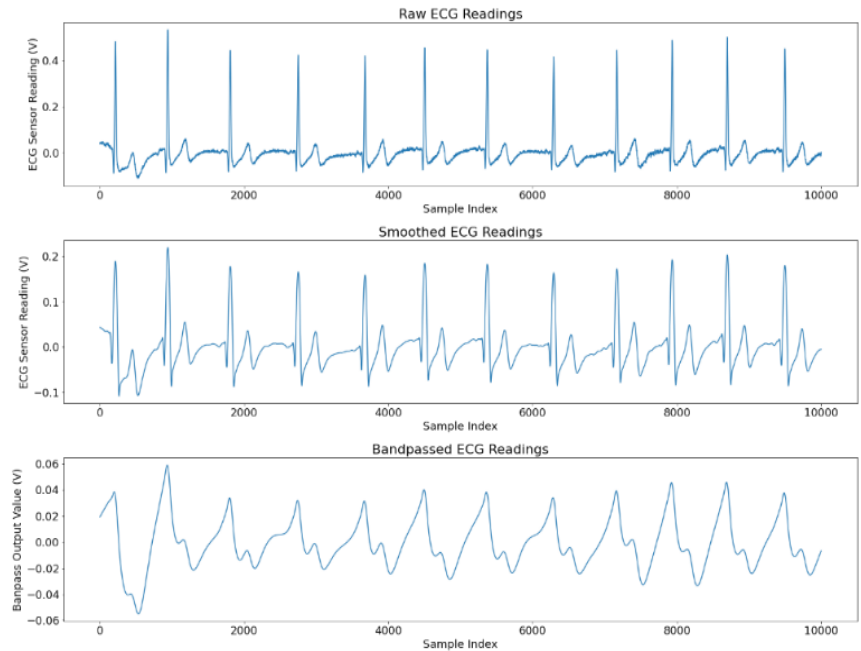

Figure 5. Raw acquired ECG signal (Top), smoothed ECG signal using Savitzky–Golay filter (Middle), and bandpass filter output (Bottom)

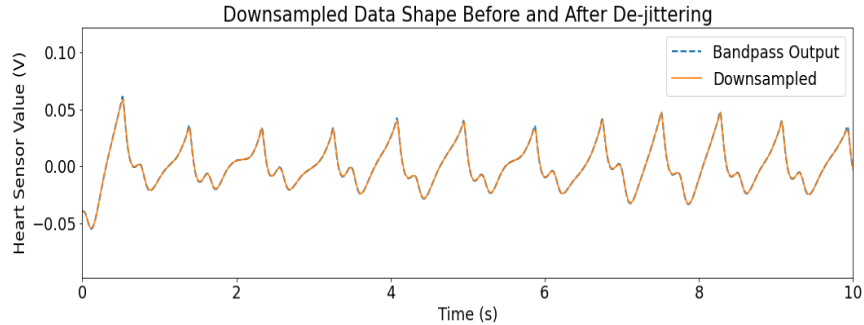

Figure 6. Comparison of ECG data shape before and after downsampling to camera's frame rate.

## 5. CONCLUSION

In this work, we propose a novel technique for removing jitter from sensor and camera timestamps and describe the effort and fundamental tasks involved in preparing visual heart rate data to be used by machine learning and image analysis techniques. The proposed pipeline starts with the removal of any unusable data, followed by timestamp correction, signal analysis and processing, and finally frame annotation. And while our de-jittering method has been presented within the context of heart rate signal estimation, it can be used to correct jitter in any acquisition timestamp provided that the sensor sampling rate is known.

Our proposed pipeline ensures that timestamps from both the visual sensor and the heart pulse sensor are accurately de-jittered and aligned to ensure accurate pulse signal labelling for each frame. Depending on the method being developed, further steps could be undertaken, or a different algorithm could be used. For example, some deep learning models take differences between consecutive frames as input [12]. However, the steps we outline in this paper usually find their place in most dataset preparation pipelines even when not explicitly mentioned with the published results.

## Acknowledgment

The research conducted in this publication was funded by the Irish Research Council under project ID EBPPG/2021/92 as a part of the Employment-Based Programme Postgraduate Scholarship in partnership with the Xperi Corporation.